# Clustering and Bayesian network for image of faces classification

Khlifia Jayech 1
SID Laboratory, National Engineering School of Sousse
Technology Park 4054 Sahloul, Sousse
Tunisia
jayech_k@yahoo.fr

Mohamed Ali Mahjoub 2
Preparatory Institute of Engineer of Monastir
Street Ibn Eljazzar Monastir
Tunisia
Medali.mahjoub@ipeim.rnu.tn

*Abstract*— In a content based image classification system, target images are sorted by feature similarities with respect to the query (CBIR). In this paper, we propose to use new approach combining distance tangent, k-means algorithm and Bayesian network for image classification. First, we use the technique of tangent distance to calculate several tangent spaces representing the same image. The objective is to reduce the error in the classification phase. Second, we cut the image in a whole of blocks. For each block, we compute a vector of descriptors. Then, we use K-means to cluster the low-level features including color and texture information to build a vector of labels for each image. Finally, we apply five variants of Bayesian networks classifiers (Naïve Bayes, Global Tree Augmented Naïve Bayes (GTAN), Global Forest Augmented Naïve Bayes (GFAN), Tree Augmented Naïve Bayes for each class (TAN), and Forest Augmented Naïve Bayes for each class (FAN) to classify the image of faces using the vector of labels. In order to validate the feasibility and effectively, we compare the results of GFAN to FAN and to the others classifiers (NB, GTAN, TAN). The results demonstrate FAN outperforms than GFAN, NB, GTAN and TAN in the overall classification accuracy.

*Keywords; face recognition, clustering, Bayesian network, Naïve Bayes, TAN, FAN*

## I. Introduction

Classification is a basic task in data mining and pattern recognition that requires the construction of a classifier, that is, a function that assigns a class label to instances described by a set of features (or attributes) [15]. Learning accurate classifiers from pre-classified data has been a very active research topic in the machine learning. In recent years, numerous approaches have been proposed in face recognition for classification such as Fuzzy sets, Rough sets, Hidden Markov Model (HMM), Neural Network, Support Vector Machine and Genetic Algorithms, Ant Behavior Simulation, Case-based Reasoning, Bayesian Networks etc. Much of the related work on image classification for indexing, classifying and retrieval has focused on the definition of low-level descriptors and the generation of metrics in the descriptor space [2]. These descriptors are extremely useful in some generic image classification tasks or when classification based on query by example. However, if the aim is to classify the image using the descriptors of the object content this image.

Several methods have been proposed for face recognition and classification, we quote: structural methods and global techniques. Structural techniques deal with local or analytical characteristics. It is to extract the geometric and structural features that constitute the local structure of face of image. Thus, analysis of the human face is achieved by the individual description of its different parts (eyes, nose, mouth, ..) and by measuring the relative positions of one by another. The class of global methods includes methods that enhance the overall properties of the face. Among the most important approaches, we quote Correlation Technique used by Alexandre [Lemieux 03] is based on a simple comparison between a test image and face learning, Principal component analysis approach (eigenfaces) based on principal component analysis (PCA) [52-53], discrete cosine transform technique (DCT) which based on computing the discrete cosine transform [58], Technique using Neural Networks and support vector machine (SVM*)* In [45].

There are two questions to be answered in order to solve difficulties that are hampering the progress of research in this direction. Firstly, how to link semantically objects in images with high-level features? That's mean how to learn the dependence between objects that reflected better the data? Secondly, how to classify the image using the structure of dependence finding? Our paper presents a work which uses three variants of naïve Bayesian Networks to classify image of faces using the structure of dependence finding between objects.

This paper is divided as follows:

Section 2, presents an overview of distance tangent; Section 3 describes the developed approach based in Naïve Bayesian Network, we describe how the feature space is extracted; and we introduce the method of building the Naïve Bayesian network, Global Tree Augmented Naïve Bayes (GTAN), Global Forest Augmented Naïve Bayes (GFAN), Tree Augmented Naïve Bayes (TAN) and Forest Augmented Naïve Bayes (FAN)and inferring posterior probabilities out of the network; Section 4 presents some experiments; finally, Section 5 presents the discussion and conclusions.

## II. TANGENT DISTANCE

The tangent distance is a mathematical tool that can compare two images taking into account small transformations (rotations, translations, etc.).. Introduced in the early 90s by Simard [60] it was combined with different classifiers for character recognition, detection and recognition of faces and recognition of speech. It is still not widely used. The distance of an image to another image I1 I2 is calculated by measuring the distance between the parameter spaces via I1 and I2 respectively. These spaces locally model all forms generated by the possible transformations between two images.

When an image $x \in R^{I \times J}$ is transformed (e.g. scaled and rotated) by a transformation $t(x, \alpha)$ which depends on $L$ parameters $\alpha \in R^L$ (e.g. the scaling factor and rotation angle), the set of all transformed patterns

$$M_x = \{t(x, \alpha): \alpha \in R^L\} \in R^{I \times J} \quad (1)$$

is a manifold of at most dimension $L$ in pattern space. The distance between two patterns can now be defined as the minimum distance between their respective manifolds, being truly invariant with respect to the $L$ regarded transformations. Unfortunately, computation of this distance is a hard non-linear optimization problem and the manifolds concerned generally do not have an analytic expression. Therefore, small transformations of the pattern $x$ are approximated by a tangent subspace $\widehat{M_x}$ to the manifold $M_x$ at the point $x$. This subspace is obtained by adding to $x$ a linear combination of the vectors $(x), l = 1, \dots, L$ that span the tangent subspace and are the partial derivatives of $t(x, \alpha)$ with respect to $\alpha_l$. We obtain a first-order approximation of $M_x$

$$M_x = \{x + \sum_{l=1}^{L} \alpha_l T_l(x) : \alpha \in R^L\} \in R^{I \times J}$$

The single-sided (SS) TD is defined as :

$$D_{DT}(x, \mu) = \min_\alpha \{ ||x + \sum_{l=1}^{L} \alpha_l T_l(x) - \mu|| \}$$

The tangent vectors $T_l(x)$ can be computed using finite differences between the original image $x$ and a reasonably small transformation of $x$. Example images that were computed using 3 are shown in Figure 1(with the original image on the left).

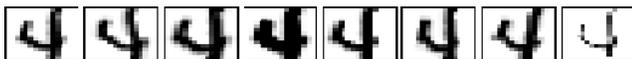

**Figure 1:** Examples for tangent approximation
.

## III. BAYESIAN NETWORK

### A. Definition

Bayesian networks represent a set of variables in the form of nodes on a directed acyclic graph (DAG). It maps the conditional independencies of these variables. Bayesian networks bring us four advantages as a data modeling tool [20-22-24]. Firstly, Bayesian networks are able to handle incomplete or noisy data which is very frequently in image analysis. Secondly, Bayesian networks are able to ascertain causal relationships through conditional independencies, allowing the modeling of relationships between variables. The last advantage is that Bayesian networks are able to incorporate existing knowledge, or pre-known data into its learning, allowing more accurate results by using what we already know.

Bayesian network is defined by :

- A directed acyclic graph (DAG) G= (V, E), where V is a set of nodes of G, and E of of G ;
- A finite probabilistic space $(\Omega, Z, p)$ ;
- A set of random variables associated with graph nodes and defined on $(\Omega, Z, p)$ as :

$$p(V_1, V_2, \dots, V_n) = \prod_{i=1}^{n} p(V_i | C(V_i))$$

where $C(V_i)$ is a set causes (parents) $V_i$ in graph G.

### B. Inference in bayesian network

Suppose we have a Bayesian network defined by a graph and the probability distribution associated with (G, P). Suppose that the graph is composed of n nodes, denoted $X = \{X_1, X_2, \dots, X_n\}$. The general problem of inference is to compute $p(X_i|Y)$ where $Y \subset X$ and $X_i \notin Y$. To calculate these conditional probabilities we can use methods of exact or approximate inferences. The first gives an accurate result, but is extremely costly in time and memory. The second turn, requires less resources but the result is an approximation of the exact solution.

To calculate these conditional probabilities we can use methods of exact or approximate inferences. The first gives an accurate result, but is extremely costly in time and memory. The second turn, requires less resources but the result is an approximation of the exact solution. A BN is usually transformed into a decomposable Markov network [59] for inference. During this transformation, two graphical operations are performed on the DAG of a BN, namely, moralization and triangulation.

### C. Parameters learning

In this case the structure is completely known a priori and all variables are observable from the data, the learning of conditional probabilities associated with variables (network nodes) may be from either a randomly or according to a Bayesian approach. The statistical learning calculation value frequencies in the training data is based on the maximum likelihood (ML) defined as follows:

$$\hat{p}(Xi = x_k \mid pa(Xi) = x_j) = \hat{\theta}_{i,j,k}^{MV} = \frac{N_{i,j,k}}{\sum_k N_{i,j,k}}$$

where $N_{i,j,k}$ is the number of events in the data base for which the variable $x_i$ in $x_k$ the state of his parents are in the configuration $x_j$.

The Bayesian approach for learning from complete data consists to find the most likely θ given the data observed using the method of maximum a posteriori (MAP) where :

$$\theta^{\wedge MAP} = argmax\ p(\theta|D) = argmax\ p(D|\theta)p(\theta)$$

With the conjugate prior distribution ; $p(\theta)$ est la distribution de Dirichlet :

$$p(\theta) = \prod_{i=1}^{n}\prod_{j=1}^{qi}\prod_{k=1}^{ri}(\theta_{i,j,k})^{\alpha_{i,j,k}-1}$$

And the posterior parameter $(D|\theta)$ :

$$p(\theta|D) = \prod_{i=1}^{n}\prod_{j=1}^{qi}\prod_{k=1}^{ri}(\theta_{i,j,k})^{N_{i,j,k}+\alpha_{i,j,k}-1}$$

Thus

$$\hat{p}(Xi = x_k \mid pa(Xi) = x_j) = \hat{\theta}_{i,j,k}^{MAP} = \frac{N_{i,j,k} + \alpha_{i,j,k} - 1}{\sum_k(N_{i,j,k} + \alpha_{i,j,k} - 1)}$$

*D. Structure learning*

Structure learning is the act of finding a plausible structure for a graph based on data input. However, it has been proven that this is an NP-Hard problem, and therefore any learning algorithm that would be appropriate for use on such a large dataset such as microarray data would require some form of modification for it to be feasible. It is explained by Spirtes *et al* (2001) that finding the most appropriate DAG from sample data is large problem as the number of possible DAGs grows super-exponentially with the number of nodes present. The number of possible structures is super-exponential in the number of variables. The number of possible combinations G of DAGs of *n* variables can be calculated by the recursive formula [20]

$$G(n) = \sum_{k=1}^{n}(-1)^{k+1}\binom{n}{k}2^{k(n-k)}G(n-k).$$

In practice we use heuristics and approximation methods like K2 algorithm.

*E. Bayesian network as a classifier*

*1)    Naïve bayes*

A variant of Bayesian Network is called Naïve Bayes. Naïve Bayes is one of the most effective and efficient classification algorithms.

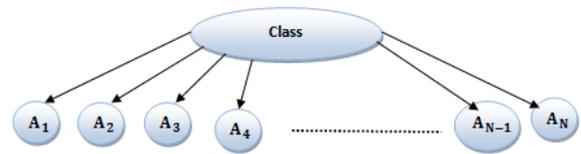

Figure 2: Naïve Bayes structure

The conditional independence assumption in naïve Bayes is rarely true in reality. Indeed, naive Bayes has been found to work poorly for regression problems (Frank et al., 2000), and produces poor probability estimates (Bennett, 2000). One way to alleviate the conditional independence assumption is to extend the structure of naive Bayes to represent explicitly attribute dependencies by adding arcs between attributes.

*2)    TAN*

An extended tree-like naive Bayes (called Tree augmented naive Bayes (TAN) was developed, in which the class node directly points to all attribute nodes and an attribute node can have only one parent from another attribute node (in addition to the class node). Figure 3 shows an example of TAN.

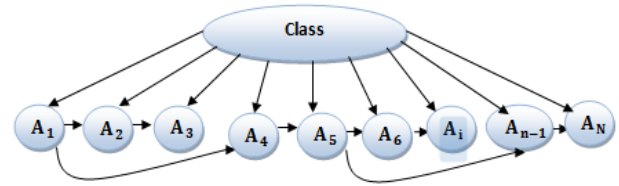

Figure 3: Tree Augmented Naïve Bayes structure (example)

Construct-TAN procedure is described in our previous work [5].

*3)    FAN*

To improve the performance of TAN, we have added a premature stop criterion based on a minimum value of increase in the score in the search algorithms of maximum weight spanning tree. (For more details view [5]).

IV.    PROPOSED SYSTEM

In this section, we present two architectures of the classification system developed face. We recall the architecture developed in our article [5] and the new architecture developed in this work. In the latter, we proposed a Bayesian network for each class. So we all structures as classes in the training set. Each structure models the dependencies between different objects in the face image. The proposed system comprises three main modules: a module for extracting primitive blocks from the local cutting of facial images, a classification module of primitives in the cluster by using the method of k-means, and a classification module the overall picture based on the structures of Bayesian networks developed for each class.

The developed system for faces classification is shown in the following figure:

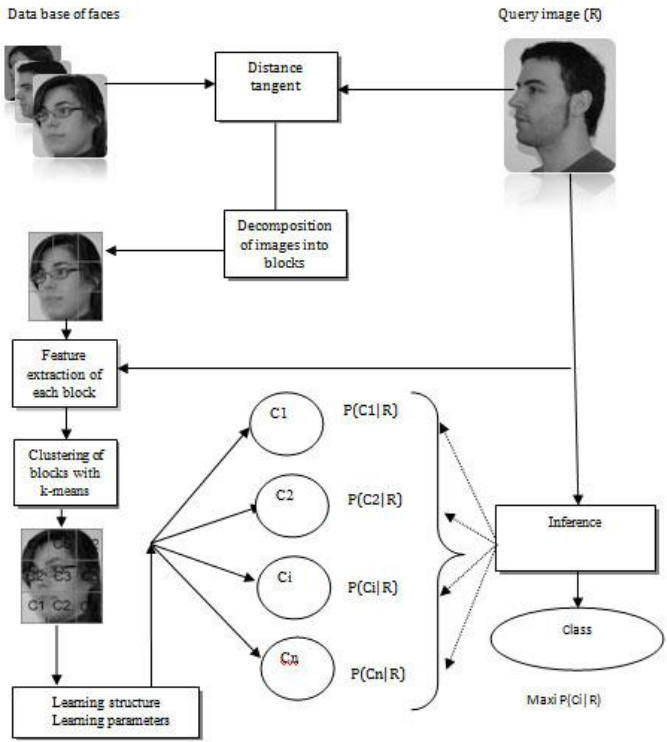

Figure 4 : Architecture Globale du Système

Emphasize at this level we will adopt two types of Bayesian networks. On the one hand, a global network (figure 5) for all classes and other one network for every class.

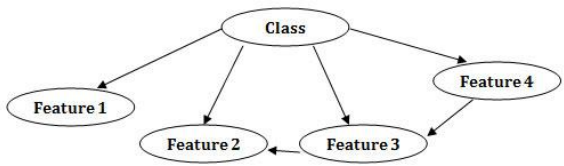

Figure 5: A Bayesian network for one class. A network such as this is created for each class

This sub-section describes the different step of the proposed system:

*A.    Step 1: decomposition of images into blocks*

*B.    Step 2: Features Extraction*

The feature vector was constructed on the basis of the average (Eq. 1), the standard deviation σ (equation 2) calculated from model mixtures of Gaussians, and energy (equation 3), the entropy (Equation 4), contrast (equation 5) and homogeneity (equation 6) derived from the co-occurrence matrix of gray levels used to cut blocks of the image. These characteristics are very used in the synthesis of texture. The average gives the average value (or moderate) levels of gray from all the pixels in the region of interest (ROI). This parameter represents the location of the histogram on the scale of grays. The images that have a higher average appear brighter. The standard deviation expresses the clustering (or dispersion) around the mean value. Values are more scattered, the greater the standard deviation is large. Most values are clustered around the mean, minus the standard deviation is high. Energy is a parameter that measures the homogeneity of the image. The energy has a value far lower than a few homogeneous areas. In this case, there are a lot of gray level transitions. Homogeneity is a parameter that has a behavior opposite contrast. Has more texture and more homogeneous regions of the parameter is high.

The mean, standard deviation, energy, entropy, contrast, and homogeneity are computed as follow:
The means is defined as:
$$\mu_{mn} = \frac{\sum_{x=1}^{M}\sum_{y=1}^{N} I_i(x,y)}{M \times N} \quad (1)$$
The standard deviation is defined as:
$$\sigma^2 = \frac{\sum_{x=1}^{M}\sum_{y=1}^{N}[I_i(x,y)-\mu]^2}{M \times N} \quad (2)$$
The energy is defined as:
$$E = \sum_{i,j}(p(i,j))^2 \quad (3)$$
The entropy is defined as
$$ENT = -\sum_i \sum_j p(i,j) \log p(i,j) \quad (4)$$
The contrast is defined as
$$CONT = \sum_i \sum_j (i-j)^2 p(i,j) \quad (5)$$
The homogeneity is defined as
$$HOM = \sum_{i,j} \frac{1}{1+(i-j)^2} p(i,j) \quad (6)$$

*C.    Step 3: Clustering of blocs with K-means:*

Our aproach is based on modeling of the image by a 3x3 grid reflecting a local description of the image. So the image to be processed is considered a grid of nine blocks (figure 6). At each block we will apply the descriptors presented in the previous section.

To generate the label vector from vector descriptor, we used the k-means algorithm. Each vector will undergo a clustering attribute, and replace the labels generated vector components descriptors. We use the method of k-means to cluster the descriptor as shown in figure?

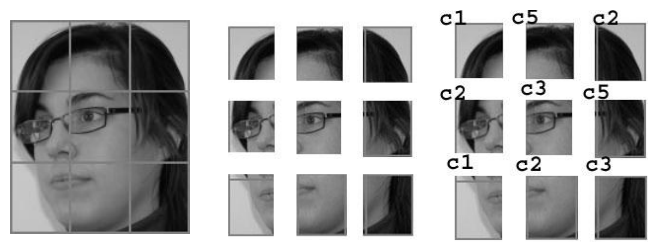

Figure 6: The result of clustering with k-means into five clusters

### D. Step 4: Structure Learning

The originality of this work is the development of a Bayesian network for each class. Then, we have compared the results of this network to a global Bayesian network. We have utilized naïve Bayes, Global Structure Tree Augmented Naïve Bayes, Global Structure Forest Augmented Naïve Bayes, Tree Augmented Naïve Bayes for each class (TAN), and Forest Augmented Naïve Bayes for each class (FAN) classifiers. This sub-section describes the implementation of these methods

### E. Step 5: Parameters learning

NB, TAN and FAN classifiers parameters were obtained by using the procedure as follows.

In implementation of NB, TAN and FAN, we used the Laplace estimation to avoid the zero-frequency problem. More precisely, we estimated the probabilities $P(c), P(a_i|c)$ and $P(a_i|a_j, c)$ using Laplace estimation as follows :

$$P(c) = \frac{N(c) + 1}{N + k} \quad P(a_i|c) = \frac{N(c, a_i) + 1}{N(c) + v_i}$$

$$P(a_i|a_j, c) = \frac{N(c, a_i, a_j) + 1}{N(c, a_j) + v_i}$$

Where -   N: is the total number of training instances.
- k: is the number of classes,
- $v_i$: is the number of values of attribute $A_i$,
- $N(c)$: is the number of instances in class c,
- $N(c, a_i)$: is the number of instances in class c and with $A_i = a_i$,
- $N(c, a_j)$: is the number of instances in class c and with $A_j = a_j$,
- $N(c, a_i, a_j)$: is the number of instances in class c and with $A_i = a_i$ and $A_j = a_j$.

### F. Step 6: Classification

In this work the decisions are inferred using Bayesian Networks. Class of an example is decided by calculating posterior probabilities of classes using Bayes rule.  This is described for both classifiers.

➢ **NB classifier**

In NB classifier, class variable maximizing equation (7) is assigned to a given example.
$$P(C|A) = P(C)P(A|C) = P(C) \prod_{i=1}^{n} P(A_i|C) \quad (7)$$

➢ **TAN and FAN classifiers**

In TAN and FAN classifiers, the class probability P (C|A) is estimated by the following equation defined as:

$$P(C|A) = P(C) \prod_{i=1}^{n} P(A_i|A_j, C)$$

Where $A_j$ is the parent of $A_i$ and

$$\begin{cases} P(A_i|A_j, C) = \frac{N(c, a_i, a_j)}{N(c, a_j)} & \text{si } A_j \text{ existe} \\ P(A_i|A_j, C) = \frac{N(c, a_i)}{N(c)} & \text{si } A_j \text{ n'existe pas} \end{cases}$$

The classification criterion used is the most common maximum a posteriori (MAP) in Bayesian Classification problems. It is given by:

$$d(A) = \text{argmax}_{classe} P(classe|A)$$
$$= \text{argmax}_{classe} P(A|classe) \times P(classe)$$
$$= \text{argmax}_{classe} \prod_{i}^{N} P(A_i|classe) \times P(classe)$$

## V. EXPERIMENTS AND RESULTS

In this section, we present the results obtained using a database of images. We start by presenting the database with which we conducted our tests, then we present our results according to the used structure (global or one for each class)..

### A. Data base prensentation

Now, we present the results of the contribution of our approach to classify images of some examples of classes from the database used 'Database of Faces'. We have used the GTAV face database found at [1] and the ORL (figure 7) face database found at [2]. The databases are used to evaluate systems classification in the presence of facial variations in lighting conditions, facial expressions (smiling, open/closed eyes) and facial details (glasses / no glasses).

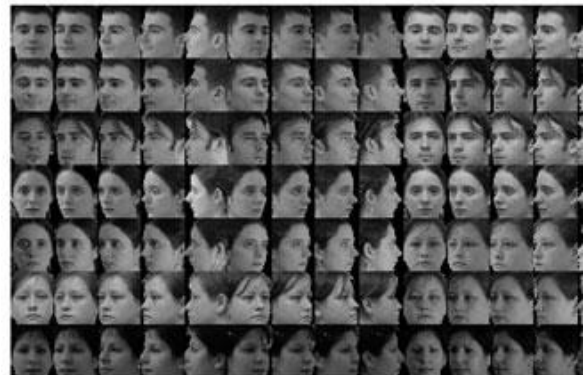

Figure 7: Example of classes of database

### B. Structure learning

We have used Matlab, and more exactly Bayes Net Toolbox of Murphy (Murphy, on 2004) and Structure Learning Package described in (Leray and al. 2004) to learn structure. Indeed, by applying the two architectures developed and the algorithm of

---

[1] http://gpstsc.upc.es/GTAV/ResearchAreas/UPCFaceDatabase/GTAVFaceDatabase.htm

[2] http://www.cl.cam.ac.uk/research/dtg/attarchive/facedatabase.html

TAN and FAN detailed in our previous work [5], we obtained the structures as follows :

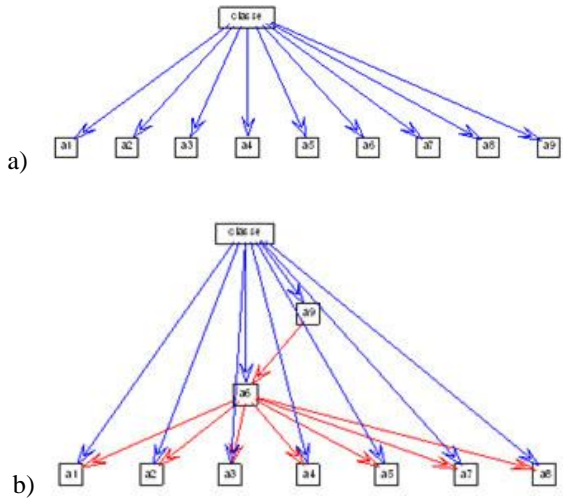

a)

b)

Figure 8 : Global network structure
a) structure of TAN b) structure of FAN

The application of structure learning algorithm based on global image and on the basis of images from each class gave us the following results:
- The structure of GFAN (Global FAN) returns to the structure of NB, that is to say, the attributes are independent of each other. In fact, the variability of face images and the choice of a very high mutual information will decrease the degree of dependency between attributes.
- On the other hand, we note that there is a dependency between the attributes of face images in the same class as shows the structures of class 1, 2, 3.4, and 5.
- By comparing the structure with those of FAN TAN, we see that the choice of a very high dependency criterion will eliminate weak dependencies between attributes and keeps only the bonds that represent a strong dependency between attributes and which will influence positively on the classification results.

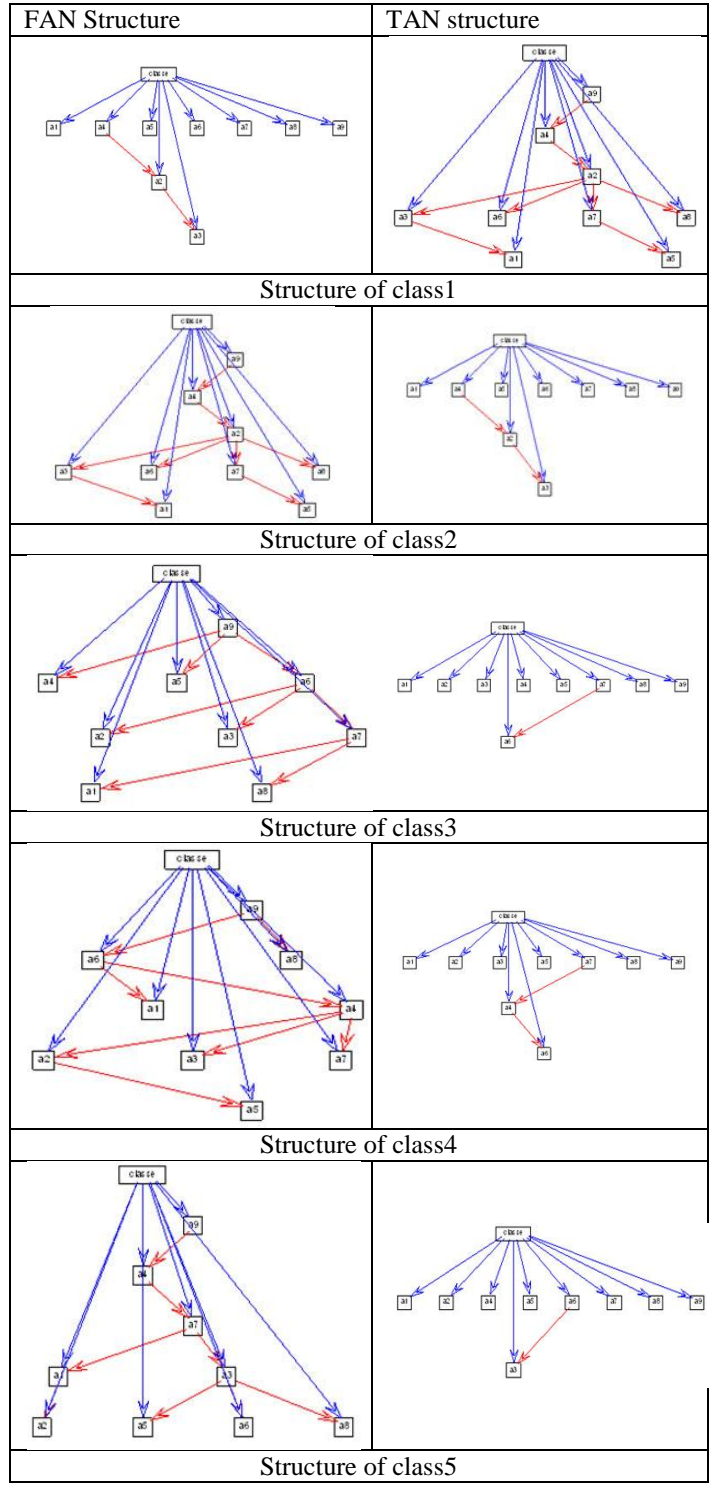

| FAN Structure | TAN structure |
|---|---|
| Structure of class1 | |
| Structure of class2 | |
| Structure of class3 | |
| Structure of class4 | |
| Structure of class5 | |

Figure 9: Bayesian Network structure for each class

*C. Parameter learning*

After estimating the global structure of GTAN, GFAN and the structure of TAN and FAN of each class. We have used those structures to estimate the conditional and a priori probability $P(c), P(F_i|c)$ and $P(F_i|F_j, c)$ using Laplace estimation to avoid the zero-frequency problem; we obtained the results as follows (Tables I, II):

TABLE I. A PRIORI PROBABILITY OF ATTRIBUTE CLASS P(CLASS)

|         | P (Class) |
|---------|-----------|
| Class 1 | 0.2       |
| Class 2 | 0.2       |
| Class 3 | 0.2       |
| Class 4 | 0.2       |
| Class 5 | 0.2       |

TABLE II. A PRIORI PROBABILITY OF $P(\text{Feature}_1|\text{CLASS})$ WITH NUMBER OF CLUSTER K=5

|           | P (Feature\|Class) ||||||
|-----------|---------|---------|---------|---------|---------|
|           | Class 1 | Class 2 | Class 3 | Class 4 | Class 5 |
| Feature 1 | 0,80    | 0,84    | 0,28    | 0,84    | 0,76    |
| Feature 2 | 0,04    | 0,04    | 0,04    | 0,04    | 0,04    |
| Feature 3 | 0,04    | 0,04    | 0,04    | 0,04    | 0,04    |
| Feature 4 | 0,08    | 0,04    | 0,60    | 0,04    | 0,12    |
| Feature 5 | 0,04    | 0,04    | 0,04    | 0,04    | 0,04    |

*D. Results*

For each experiment, we used the percentage of correct classification (PCC) to evaluate the classification accuracy defined as:

$$PCC = \frac{\text{number of images correctly classified}}{\text{total number of images classified}}$$

The results of experiments are summarized in Tables IV,V and figure 10 with number of cluster k=8[3], Naive Bayes, Global TAN, Global FAN, TAN and FAN use the same training set and are exactly evaluated on the same test set.

*E. Naïve Bayes*

From the table III, we note that the naive Bayesian networks, despite their simple construction gave a very good classification rates. However, this rate is improving as we see that the classification rates obtained by class 3 is 40%. So we conclude that there is a slight dependency between the attributes that must be determined by performing a structure learning.

---
[3] The choice of k value discussed in our previous work [5]

TABLE III. NAÏVE BAYES CLASSIFICATION RATES

| Network | class | Number of cluster k | Training data's classification rate | Test data's classification rate |
|---------|-------|---------------------|-------------------------------------|--------------------------------|
| Naïve Bayes | class 1 | 8 | 0,95 | 1,00 |
|  | class 2 | 8 | 1,00 | 0,90 |
|  | class 3 | 8 | 0,75 | 0,40 |
|  | class 4 | 8 | 0,90 | 0,90 |
|  | class 5 | 8 | 0,80 | 0,90 |

*F. Tree augmented Naïve Bayes*

From the table above, we note that the rate of correct classification decreases. However, there is a slight improvement in rate of correct classification using a class structure.

*G. Forest augmented Naïve Bayes*

From the table above, we find that the rate of correct classification GFAN is the same as that obtained by naive Bayes. Since learning of GFAN structure with a strict threshold equivalent to 0.8 gave the same structure as Naïve Bayes. However, using the structure of class 1, we note that the classification rate was slightly improved to class 1. Also for the Class 3 classification rate increased from 0.4 to 0.9 using the structure for class 3 FAN

*H. Discussion*

According to our experiments, we observe that the naive Bayesian network gave a good result as NAT. Two factors may cause:

➢ The directions of links are crucial in a NAT. According to the TAN algorithm detailed in our article [5] an attribute is randomly selected as the root of the tree and the directions of all links are made thereafter. We note that the selection of the root attribute actually determines the structure of the TAN result, since TAN is a directed graph. Thus the selection of the root attribute is important to build a NAT.

➢ Of unnecessary links can exist in a NAT. According to the TAN algorithm, a spanning tree of maximum weight is constructed. Thus, the number of links is set to n-1. Sometimes it could be a possible bad fit of the data, since some links may be unnecessary to exist in the NAT.

It is observed that NB and Global Fan gave the same classification rate, since they have the same structure.
We note also that the rate of correct classification given by FAN is very high that TAN. Several factors are involved:

1. According to the FAN algorithm illustrated in our article [5], the choice of the attribute A root is defined by the equation below, the maximum mutual root has the information with the class.,

$$A_{racine} = \text{argmax}_{A_i} I_p(A_i; C)$$

when $i = 1, ..., n$.. It is obvious to use this strategy, ie the attribute that has the greatest influence on the classification should be the root of the tree.

2. Filtering of links that have less than a conditional mutual information threshold. These links are at high risk for a possible bad fit of the data which could distort the calculation of conditional probabilities. Specifically, the use of a conditional average mutual information defined in the equation below as a threshold. All links that have conditional mutual information unless Iavg are removed.

$$\text{Iavg} = \frac{\sum_i \sum_{j, j \neq i} I_p(A_i; A_j | C)}{n(n-1)}$$

where n is the number of attributs.

TABLE IV. RESULTS FOR TREE AUGMENTED NAÏVE BAYES

| Network | Structure | class | Training data's classification accuracy | Test data's classification accuracy |
|---|---|---|---|---|
| GTAN | Global Structure (0,84 seconds) | Class 1 | 0,65 | 0,60 |
| | | Class 2 | 0,50 | 0,40 |
| | | Class 3 | 0,50 | 0,70 |
| | | Class 4 | 0,40 | 0,40 |
| | | Class 5 | 0,40 | 0,50 |
| TAN (Tree Augmented Naive Bayes) | Structure class 1 (0,75 seconds) | Class 1 | 0,85 | 0,65 |
| | | Class 2 | 0,60 | 0,50 |
| | | Class 3 | 0,40 | 0,40 |
| | | Class 4 | 0,50 | 0,30 |
| | | Class 5 | 0,45 | 0,40 |
| | structure class 2 (0,72 seconds) | Class 1 | 0,40 | 0,30 |
| | | Class 2 | 0,60 | 0,55 |
| | | Class 3 | 0,45 | 0,40 |
| | | Class 4 | 0,50 | 0,40 |
| | | Class 5 | 0,55 | 0,50 |
| | structure class 3 (0,81 seconds) | Class 1 | 0,45 | 0,4 |
| | | Class 2 | 0,5 | 0,4 |
| | | Class 3 | 0,55 | 0,50 |
| | | Class 4 | 0,45 | 0,30 |
| | | Class 5 | 0,40 | 0,3 |
| | structure class 4 (0,95 seconds) | Class 1 | 0,55 | 0,50 |
| | | Class 2 | 0,55 | 0,40 |
| | | Class 3 | 0,60 | 0,50 |
| | | Class 4 | 0,70 | 0,60 |
| | | Class 5 | 0,55 | 0,50 |
| | structure class 5 (0,88 seconds) | Class 1 | 0,60 | 0,50 |
| | | Class 2 | 0,45 | 0,30 |
| | | Class 3 | 0,55 | 0,40 |
| | | Class 4 | 0,50 | 0,40 |
| | | Class 5 | 0,65 | 0,60 |

TABLE V. RESULTS FOR FOREST AUGMENTED NAÏVE BAYES

| Network | Structure | class | Number of cluster k | Training data's classification accuracy | Test data's classification accuracy |
|---|---|---|---|---|---|
| GFAN | Global Structure (0,84 seconds) | Class 1 | 8 | 0,95 | 1 |
| | | Class 2 | 8 | 1 | 0,9 |
| | | Class 3 | 8 | 0,75 | 0,4 |
| | | Class 4 | 8 | 0,9 | 0,9 |
| | | Class 5 | 8 | 0,8 | 0,9 |
| FAN with threshold S=0,8 | structure of class 1 (0,75 seconds) | Class 1 | 8 | 1,00 | 1,00 |
| | | Class 2 | 8 | 1,00 | 0,80 |
| | | Class 3 | 8 | 0,70 | 0,80 |
| | | Class 4 | 8 | 0,70 | 0,70 |
| | | Class 5 | 8 | 0,75 | 0,80 |
| | structure of class 2 (0,72 seconds) | Class 1 | 8 | 0,85 | 0,80 |
| | | Class 2 | 8 | 1,00 | 0,90 |
| | | Class 3 | 8 | 0,70 | 1,00 |
| | | Class 4 | 8 | 0,75 | 0,90 |
| | | Class 5 | 8 | 0,70 | 0,90 |
| | structure of class 3 (0,81 seconds) | Class 1 | 8 | 0,80 | 0,80 |
| | | Class 2 | 8 | 0,90 | 0,70 |
| | | Class 3 | 8 | 0,90 | 0,90 |
| | | Class 4 | 8 | 0,80 | 0,90 |
| | | Class 5 | 8 | 0,80 | 0,30 |
| | structure of class 4 (0,95 seconds) | Class 1 | 8 | 0,75 | 0,80 |
| | | Class 2 | 8 | 1,00 | 0,90 |
| | | Class 3 | 8 | 0,75 | 0,80 |
| | | Class 4 | 8 | 1.00 | 0,90 |
| | | Class 5 | 8 | 0,75 | 0,70 |
| | structure of class 5 (0,88 seconds) | Class 1 | 8 | 0.90 | 0,80 |
| | | Class 2 | 8 | 0,90 | 0,90 |
| | | Class 3 | 8 | 0,75 | 0,50 |
| | | Class 4 | 8 | 0,75 | 0,80 |
| | | Class 5 | 8 | 0,95 | 0,90 |

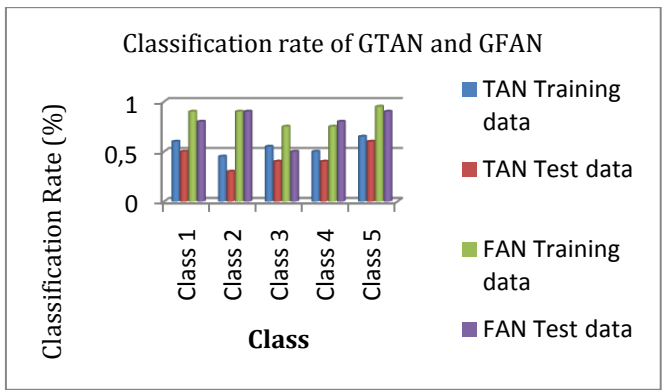

a)

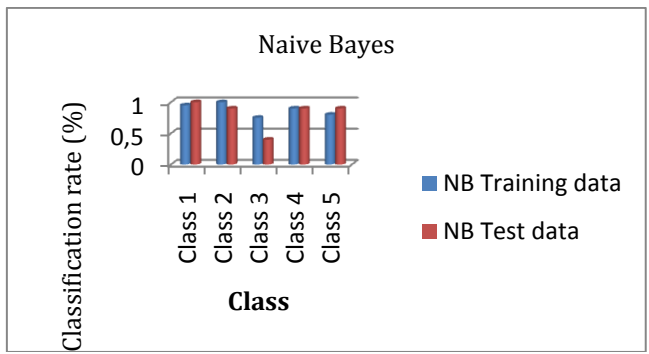

b)

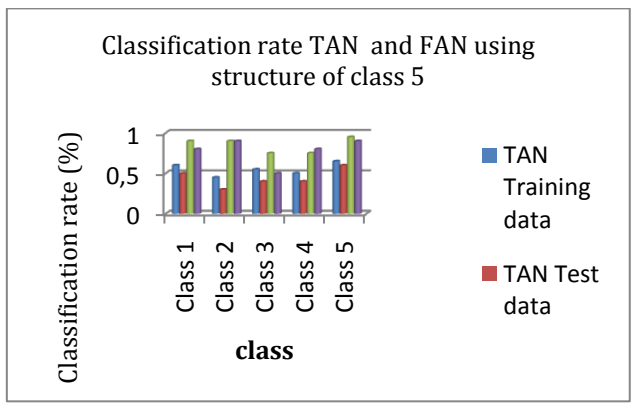

c)

Figure 10 : Classification rates
a) GTAN and GFAN b) Naive Bayes
c) TAN and FAN using structure of class 5

## VI. CONCLUSION

This study is an extension of our previous work [5]. We have developed a new approach for classifying image of faces using distance tangent and the method of k-means to cluster the vectors descripteurs of the images. First, we use the technique of tangent distance to calculate several tangent spaces representing the same image. The objective is to reduce the error in the research phase.

Then we have used Bayesian network as classifier to classify the whole image into five classes. We have implemented and compared three classifiers: NB TAN and FAN using two types of structure, a global structure and structure per class presenting respectively the dependence between the attributes inter and intra-class. The goal was to compare the results obtained by these structures and apply algorithms that can produce useful information from a high dimensional data. In particular, the aim is to improve Naïve Bayes by removing some of the unwarranted independence relations among features and hence we extend Naïve Bayes structure by implementing the Tree Augmented Naïve Bayes. Unfortunately, our experiments show that TAN performs even worse than Naïve Bayes in classification. Responding to this problem, we have modified the traditional TAN learning algorithm by implementing a novel learning algorithm, called Forest Augmented Naïve Bayes. We experimentally test our algorithm in data image of faces and compared it to NB and TAN. The experimental results show that FAN improves significantly NB classifiers' performance in classification. In addition, the results show that the mean of classification accuracy is better when the number of cluster is optimal that's mean the number of cluster that can reflected better the data. Then, we marked that the structure of FAN per class performs better than Global FAN. This results is explained by the use of structure of FAN per class reflect better the dependence of the attribute intra-class (in the same class), and the use of a global structure reflects better the dependence inter-class (between the classes).

AUTHORS PROFILE

**Khlifia Jayech** is a Ph.D. student at the department of Computer Science of National School of Computer Sciences. She obtained her master degree from Higher Institute of Applied Science and Technology of Sousse (ISSATS), in Janvier 2010. Her areas of research include Image analysis, processing, indexing and retrieval, Dynamic Bayesian Network, Hidden Markov Model and recurrent neural network.

**Dr Mohamed Ali Mahjoub** is an assistant professor in the department of computer science at the Preparatory Institute of Engineering of Monastir. His research interests concern the areas of Bayesian Network, Computer Vision, Pattern Recognition, Medical imaging, HMM, and Data Retrieval. His main results have been published in international journals and conferences.